\documentclass{article}

\PassOptionsToPackage{numbers,sort&compress}{natbib} % numeric citations sorted + merged: [2--4,7]
\usepackage[preprint]{neurips_2026}

\usepackage[utf8]{inputenc}
\usepackage[T1]{fontenc}
\usepackage{microtype}
\usepackage[table]{xcolor}     % [table] enables \rowcolor / \columncolor
\usepackage{hyperref}          % BEFORE cleveref

\usepackage{amsmath}
\usepackage{amssymb}
\usepackage{amsfonts}
\usepackage{mathtools}
\usepackage{amsthm}

\usepackage{graphicx}
\usepackage{subcaption}
\usepackage{wrapfig}           % text-wrapped figures (compact teaser layout)
\usepackage{booktabs}
\usepackage{nicefrac}
\usepackage{placeins}          % \FloatBarrier (used in 5_conclusion.tex)
\usepackage{array}             % needed for \newcolumntype
\usepackage{bbding}             % \Checkmark, \XSolidBrush
\usepackage{tikz}              % swatches in figure captions

\usepackage[capitalize,noabbrev]{cleveref}

\newcolumntype{G}{>{\columncolor[gray]{0.9}}c}
\newcolumntype{g}{>{\columncolor[gray]{0.9}}l}

\newcommand{\myparagraph}[1]{\noindent \textbf{#1}}
\newcommand{\linhan}[1]{}      % silenced; any stray review notes compile to nothing

\theoremstyle{plain}

\theoremstyle{definition}

\theoremstyle{remark}

\title{Drive-JEPA: Video JEPA Meets Multimodal Trajectory Distillation for End-to-End Driving}

\author{%
  Linhan Wang\thanks{Work done during an internship at XPENG Motors.} \\
  Virginia Tech \\
  \texttt{linhanwang@vt.edu} \\
  \And
  Zichong Yang \\ Purdue University \\
  \And
  Chen Bai\thanks{Corresponding author.} \\ XPENG Motors \\
  \And
  Guoxiang Zhang \\ XPENG Motors \\
  \AND
  Xiaotong Liu \\ XPENG Motors \\
  \And
  Xiaoyin Zheng \\ XPENG Motors \\
  \And
  Xiao-Xiao Long \\ Nanjing University \\
  \And
  Chang-Tien Lu \\ Virginia Tech \\
  \And
  Cheng Lu \\ XPENG Motors \\
}

\begin{document}
\maketitle

\begin{abstract}
End-to-end autonomous driving increasingly leverages self-supervised video pretraining to learn transferable planning representations. However, pretraining video world models for scene understanding has so far brought only limited improvements. This limitation is compounded by the inherent ambiguity of driving: each scene typically provides only a single human trajectory, making it difficult to learn multimodal behaviors. In this work, we propose Drive-JEPA, a framework that integrates Video Joint-Embedding Predictive Architecture (V-JEPA) with multimodal trajectory distillation for end-to-end driving. First, we adapt V-JEPA for end-to-end driving, pretraining a ViT encoder on large-scale driving videos to produce predictive representations aligned with trajectory planning. Second, we introduce a proposal-centric planner that distills diverse simulator-generated trajectories alongside human trajectories, with a momentum-aware selection mechanism to promote stable and safe behavior. When evaluated on NAVSIM, the V-JEPA representation combined with a simple transformer-based decoder outperforms prior methods by 3 PDMS in the perception-free setting. The complete Drive-JEPA framework achieves 93.7 PDMS on v1 and 87.8 EPDMS on v2, setting a new state-of-the-art. Code is available at \url{https://github.com/linhanwang/Drive-JEPA}.
\end{abstract}

\section{Introduction} \label{sec:intro}

End-to-end autonomous driving \cite{pomerleau1988alvinn, chitta2022transfuser, hu2023planning} has emerged as a promising paradigm that directly maps raw sensor observations to driving actions using a unified neural model. By eliminating hand-designed intermediate representations used in traditional modular pipelines, end-to-end approaches aim to reduce information loss and improve scalability by learning directly from large collections of human driving data.

\begin{wrapfigure}{r}{0.45\textwidth}
  \vspace{-12pt}
  \centering
  \includegraphics[width=0.45\textwidth]{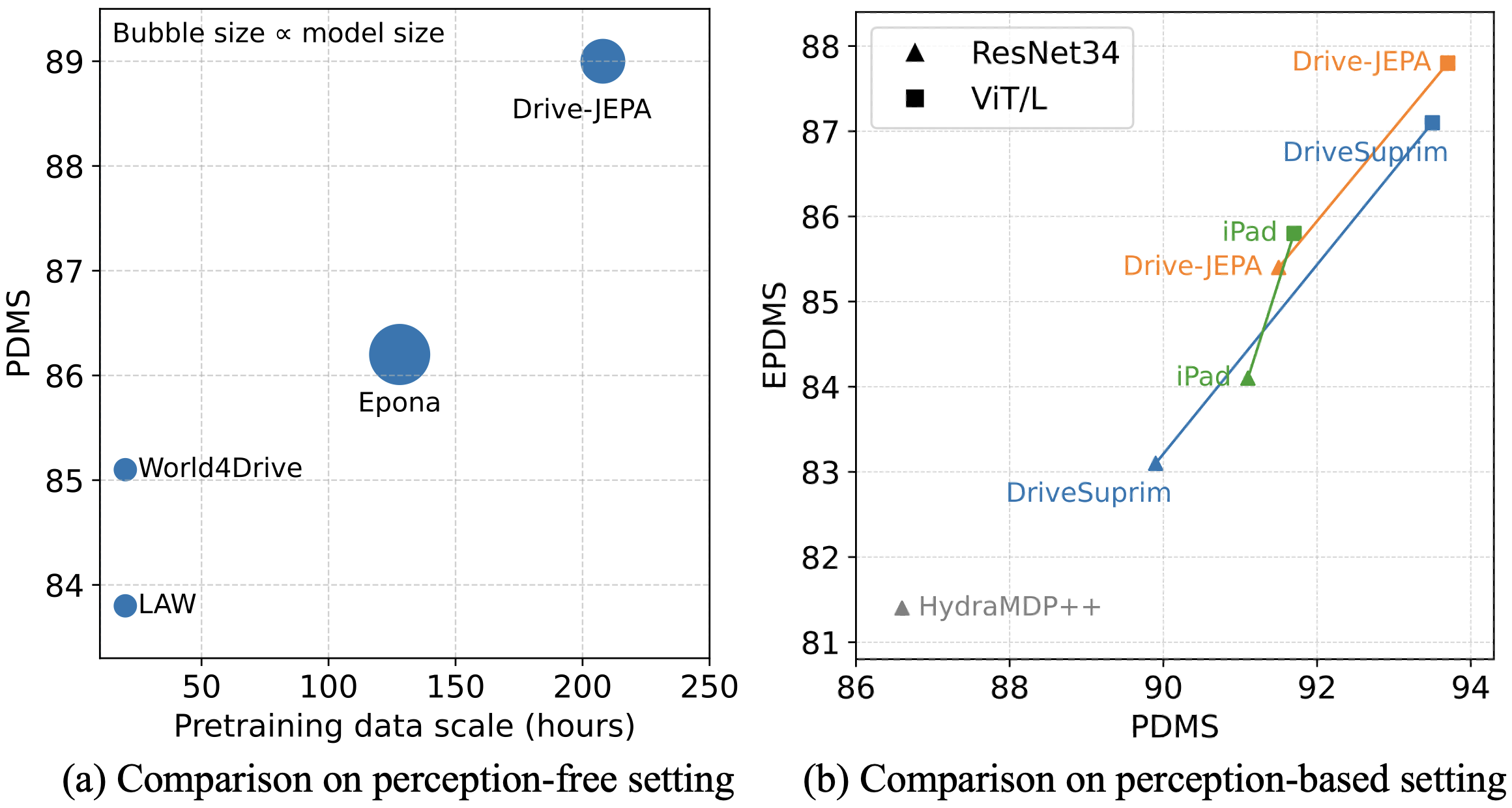}
  \caption{Comparison between end-to-end planners on both perception-free and perception-based settings.}
  \label{fig:small_teaser}
  \vspace{-12pt}
\end{wrapfigure}
Recently, end-to-end autonomous driving increasingly seeks to leverage self-supervised video pretraining to learn transferable representations for planning. However, pretraining video world models for scene understanding has so far brought only limited improvements. Existing approaches in this direction largely fall into two categories. First, video-generative methods, such as VaVAM \cite{bartoccioni2025vavim} and Epona \cite{zhang2025epona}, learn representations by reconstructing or generating videos and then transfer them to planning, but this pixel-level objective incurs heavy computation and may over-emphasize visual details that are irrelevant to decision making. Second, to reduce cost, latent world models predict compact feature dynamics (e.g., LAW \cite{lienhancing} predicts feature $T{+}1$ from feature $T$, and World4Drive \cite{zheng2025world4drive} further introduces pretrained foundation models to enrich latent targets. However, these latent approaches are typically used as auxiliary objectives and have not demonstrated clear benefits from scaling up pretraining. 

Orthogonally, end-to-end driving faces a supervision bottleneck: each scene typically provides only a single human trajectory, despite inherently multimodal futures. Prior works address this by generating multimodal trajectories through either discrete or continuous formulations. Discrete approaches such as VAD v2 \cite{chen2024vadv2} and Hydra-MDP \cite{li2024hydra} cluster trajectories into a fixed vocabulary and predict scores reflecting safety and comfort; however, their expressiveness is fundamentally limited by the coverage and quality of anchor trajectories, leading to poor generalization in out-of-vocabulary scenarios \cite{liao2025diffusiondrive}. Alternatively, diffusion-based methods, including DiffusionDrive \cite{liao2025diffusiondrive} and GoalFlow \cite{xing2025goalflow}, model multimodal trajectory distributions via iterative sampling, which has shown strong generative capability. Nevertheless, these approaches remain constrained by supervision from single human trajectories per scene, inherently limiting the diversity of learned behaviors.

%% P4: Drive-JEPA overview and two solutions to the two challenges
In this work, we propose \textbf{Drive-JEPA}, an end-to-end autonomous driving framework that addresses the above two bottlenecks in a unified way. First, we adapt V-JEPA \cite{assran2023self, assran2025v} to the driving domain to learn planning-aligned predictive representations from large-scale raw videos, improving transfer beyond prior world-model pretraining. Second, we introduce multimodal trajectory distillation that distills knowledge from simulators into a proposal-centric planner, providing diverse supervision beyond single human trajectories and enabling safer multimodal decision making.

%% P5: Drive-JEPA details (pretraining, proposal generation, and selection).
Specifically, our framework consists of three components: Driving Video Pretraining, Multimodal Trajectory Distillation, and Momentum-aware Trajectory Selection. In the first module, we curate a large-scale driving video dataset and pretrain a ViT-based vision encoder using V-JEPA \cite{assran2023self, assran2025v}, which learns predictive representation by predicting future latent with effective mode collapse prevention. In the second module, the waypoint-anchored proposals generation leverages deformable attention \cite{xia2022vision, guo2025ipad} to aggregate BEV features \cite{li2024bevformer} at trajectory waypoints and refine proposals iteratively. To increase diversity, proposals are supervised using both human trajectories and simulator-generated multimodal trajectories that satisfy safety and comfort constraints, enabling effective knowledge distillation from the simulator. Finally, the selection module assigns scores to all candidates by predicting collision risk, traffic-rule compliance, and comfort, and further incorporates a momentum-aware penalty to reduce frame-to-frame trajectory distortion.

%% P5: Evaluation summary.
We validate Drive-JEPA on NAVSIM v1 \cite{dauner2024navsim}, NAVSIM v2 \cite{Cao2025CORL} , and Bench2Drive \cite{jia2024bench}. Drive-JEPA achieves 93.7 PDMS on NAVSIM v1 and 87.8 EPDMS on NAVSIM v2, setting a new state of the art. Notably, with only a single front-view camera and a lightweight transformer planner, our V-JEPA-pretrained model outperforms prior work by 3 PDMS in the perception-free setting, highlighting the effectiveness of V-JEPA pretraining for planning. On Bench2Drive, the Multimodal Trajectory Distillation consistently improves driving quality, demonstrating the benefit of diverse supervision for generating safe, multimodal trajectories.

Our contributions can be summarized as follows:

\setlength{\parskip}{0pt}
\setlength{\itemsep}{0pt}

\begin{itemize}
    \item We introduce V-JEPA pretraining to end-to-end autonomous driving, boosting performance in both perception-based and perception-free settings.
    \item We propose a novel multimodal trajectories supervision to distill simulator knowledge to a proposal-centric framework, generating diverse multimodal trajectories.
    \item We design a momentum-aware trajectory selection module, enhancing driving comfort.
    \item Our method achieves a new state of the art on NAVSIM v1 and NAVSIM v2. In addition, our method achieves strong performance on NAVSIM even without relying on perception annotations.
\end{itemize}
\section{Related Work}

\subsection{End-to-end autonomous driving}
Early works such as ALVINN \cite{pomerleau1988alvinn} and PilotNet \cite{bojarski2016end} leverage large-scale human driving data to learn policies that map sensor observations directly to control actions. However, these models often lack interpretability and can degrade due to issues such as causal confusion. To mitigate this, recent studies incorporate intermediate representations and auxiliary supervision to improve robustness. Transfuser \cite{chitta2022transfuser} fuses LiDAR and camera features in the BEV space and strengthens BEV features with BEV segmentation and 3D detection supervision. Going further, UniAD \cite{hu2023planning} unifies the full stack of driving tasks—including tracking, mapping, and motion prediction—within a single framework jointly optimized with planning. VAD \cite{jiang2023vad} explores compact vectorized scene representations for efficiency, while SparseDrive \cite{sun2025sparsedrive} proposes a query-centric sparse structure as a BEV-free alternative. DriveTransformer \cite{jiadrivetransformer} further improves efficiency by using a small set of learned queries to aggregate multi-view image features. Moreover, DiffusionDrive \cite{liao2025diffusiondrive} and GoalFlow \cite{xing2025goalflow} investigate diffusion-based end-to-end trajectory generation. A related direction, MomAD \cite{song2025momad}, exploits cross-frame momentum for temporal consistency but regenerates the trajectory via query-space topological matching, whereas we independently cast momentum as a lightweight selection term fused with a learned scorer that jointly weighs other safety metrics. Despite these advances, end-to-end driving remains challenging due to two fundamental requirements: capturing the spatiotemporal structure of complex scenes and modeling the inherently multimodal nature of driving behaviors.

\subsection{World Models for End-to-end driving}
Video world models in autonomous driving predict how scenes evolve under ego actions. Recent progress in controllable video generation \cite{gao2024vista, hu2023gaia} has enabled action-conditioned world models, suggesting their potential as learned simulators. Motivated by the idea that realistic generation implies strong dynamics understanding, several works transfer world-model knowledge to end-to-end driving. VaVIM \cite{bartoccioni2025vavim} trains a causal auto-regressive video model and extends it to generate ego trajectories, while Epona \cite{zhang2025epona} proposes a hybrid diffusion–auto-regressive predictor with a dual-stream diffusion decoder for joint video and trajectory synthesis. However, both approaches remain computationally heavy due to pixel-level reconstruction. To improve efficiency, latent world models predict future features instead of pixels \cite{zhu2026self}. LAW \cite{lienhancing} integrates latent dynamics learning into end-to-end driving and achieves strong perception-free performance, and World4Drive \cite{zheng2025world4drive} further leverages multimodal video foundation models as richer latent targets. Nonetheless, these latent approaches have not clearly demonstrated benefits from scaling to large-scale video pretraining and may suffer from representation collapse. To overcome this, we adopt V-JEPA \cite{assran2025v} as an efficient latent world model with built-in collapse prevention, enabling scalable video pretraining for end-to-end driving.

\subsection{MultiModal Trajectories Generation}

In planning tasks such as manipulation and autonomous driving, a given scenario often offers multiple action options, requiring effective multimodal modeling. Recently, VADv2 \cite{chen2024vadv2} and Hydra-MDP \cite{li2024hydra} introduce a large fixed vocabulary of trajectories by discretizing and clustering the continuous action space. For each scene, this vocabulary provides a diverse multimodal choice space for driving behaviors, and planning is performed by selecting trajectories based on predicted scores. However, this paradigm is fundamentally limited by the coverage of the vocabulary and cannot generalize to out-of-vocabulary scenes. Other works introduce diffusion models to generate multimodal trajectories. DiffusionDrive \cite{liao2025diffusiondrive} guides the diffusion process using a small fixed vocabulary, while GoalFlow \cite{xing2025goalflow} uses goal points as guidance during diffusion-based generation. Compared with computationally expensive diffusion-based methods, proposal-centric approaches such as iPad \cite{guo2025ipad} iteratively refine a set of trajectory proposals using efficient deformable attention. Compared with Hydra-MDP, iPad’s proposals can be viewed as an online-generated continuous vocabulary; however, its candidates are supervised purely by a single human trajectory per scene, which limits diversity. In contrast, our Multimodal Trajectory Distillation breaks this limitation by distilling knowledge from simulator-generated trajectories.
\section{Method} \label{sec:method}

\subsection{Preliminary}

\paragraph{End-to-end Autonomous Driving}  In the task of end-to-end autonomous driving, the objective is to estimate the future trajectory of the ego vehicle in the form of waypoints. Formally, let \( I_t = \{ I_t^1, I_t^2, \ldots, I_t^N \} \) denote the set of \( N \) surrounding multi-view images captured at time step \( t \). The model is expected to predict a sequence of waypoints \( W_t = \{ w_t^1, w_t^2, \ldots, w_t^M \} \), where each waypoint \( w_t^i = (x_t^i, y_t^i, \psi_t^i) \) represents the predicted BEV position and heading angle of the ego vehicle at time step \( t + i \). Here, \( M \) denotes the number of future positions to be predicted. In addition, the model also takes as input the ego status of the vehicle, which includes the driving command (e.g., left, forward, right), speed, and acceleration.

\paragraph{V-JEPA}
V-JEPA \cite{bardes2024revisiting} learns predictive video representations by estimating the latent representation of a target view $y$ from a masked view $x$, where a subset of spatiotemporal patches are randomly dropped. The method adopts a meta-architecture with an encoder $E_{\theta}(\cdot)$ that extracts video features and a predictor $P_{\phi}(\cdot)$ that predicts the representations at masked locations. The encoder and predictor are optimized jointly with
\begin{equation}
\min_{\theta,\phi,\Delta_y} \; \left\| P_{\phi}(\Delta_y, E_{\theta}(x)) - \mathrm{sg}(E_{\bar{\theta}}(y)) \right\|_1,
\end{equation}
where $\Delta_y$ is a learnable mask token indicating the dropped patch locations. The target branch uses a stop-gradient operator $\mathrm{sg}(\cdot)$ and an exponential moving average encoder $E_{\bar{\theta}}$ (with parameters $\bar{\theta}$) to stabilize training and avoid representation collapse. The loss is computed only on masked positions. Both $E_{\theta}(\cdot)$ and $P_{\phi}(\cdot)$ are instantiated as Vision Transformers (ViTs)~\cite{dosovitskiy2020image}.

% \paragraph{V-JEPA} The V-JEPA objective aims to predict the learned representation of a video $y$ from a masked view $x$ of that video, a view from which patches have been randomly dropped. The task meta-architecture consists of an encoder $E_{\theta}(\cdot)$, which extracts video representations, and a predictor $P_{\phi}(\cdot)$, which predicts the representation of masked video parts. The encoder and predictor are trained simultaneously using the objective

% \begin{equation}
% \min_{\theta,\phi,\Delta_y} \; \left\| P_{\phi}(\Delta_y, E_{\theta}(x)) - \mathrm{sg}(E_{\bar{\theta}}(y)) \right\|_1,
% \end{equation}

% where $\Delta_y$ is a learnable mask token that indicates the locations of the dropped patches. The loss uses a stop-gradient operation, $\mathrm{sg}(\cdot)$, and an exponential moving average, $\bar{\theta}$, of the weights $\theta$ of the encoder network to prevent representation collapse. The loss is applied only to the predictions of the masked patches. The encoder $E_{\theta}(\cdot)$ and predictor $P_{\phi}(\cdot)$ are each parameterized as a vision transformer \cite{dosovitskiy2020image}, or ViT.

% left, bottom, right, top
\begin{figure*}[t]
  \centering
  \includegraphics[width=0.85\textwidth, trim=0 2cm 7cm 0, clip]{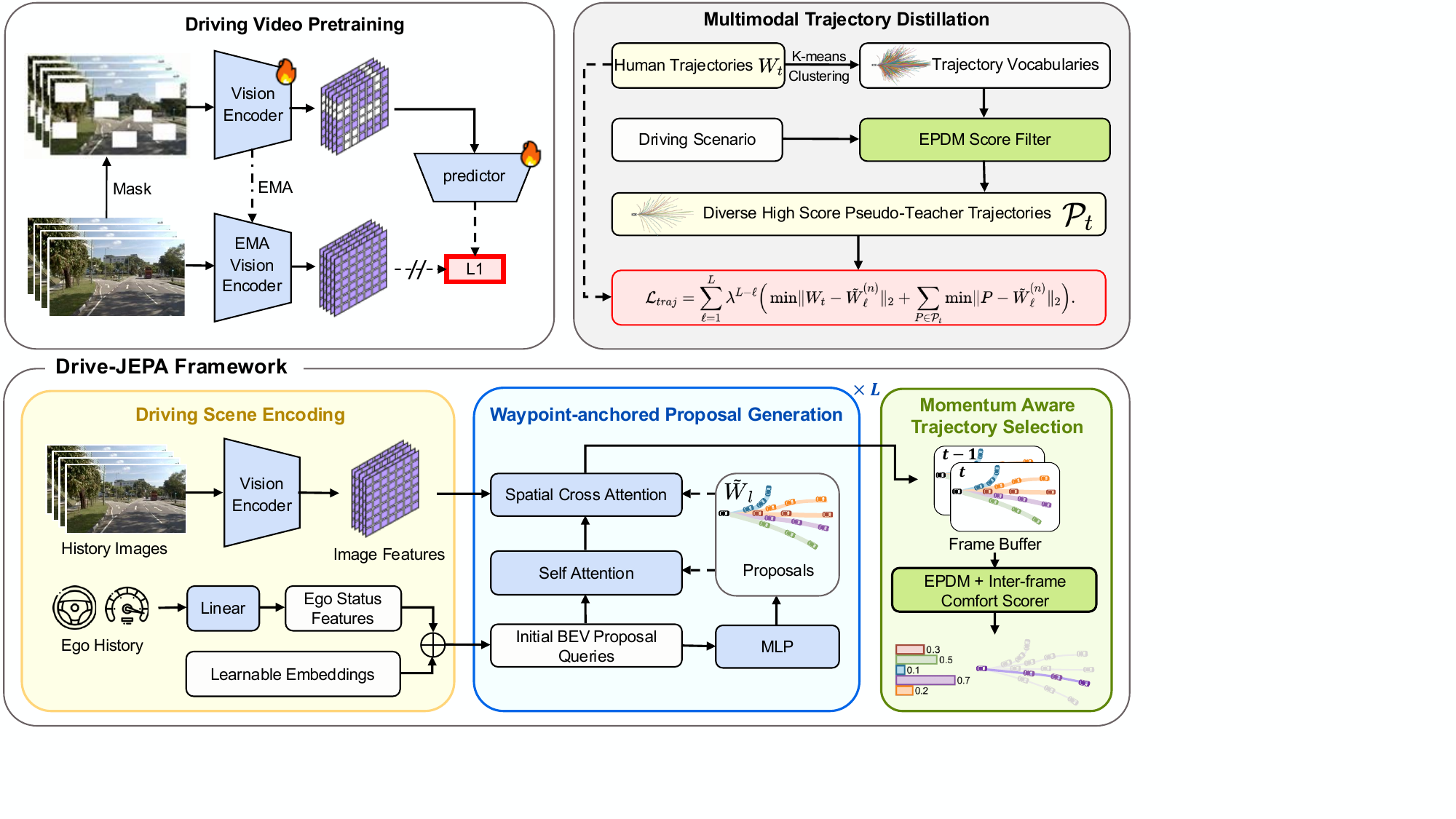}
  \caption{Overview of the Drive-JEPA architecture. Driving Video Pretraining learns a ViT encoder from large-scale driving videos using the self-supervised V-JEPA objective. Given the pretrained features, Waypoint-anchored Proposal Generation efficiently produces multiple trajectory proposals, whose distribution is guided by Multimodal Trajectory Distillation. Finally, Momentum-aware Trajectory Selection picks the final trajectory by accounting for cross-frame comfort.}
  \label{fig:structure}
\end{figure*}

\subsection{Driving Video Pretraining} \label{subsec:video_pretraining}

To enhance the representation for planning through self-supervised video pretraining, prior works explored pixel-space driving world models and latent world models. While the former faces expensive computation and the latter fails to scale, we propose to leverage V-JEPA in large-scale driving video pretraining.

\begin{wrapfigure}{r}{0.50\textwidth}
  \centering
  \scriptsize
  \vspace{-10pt}
  \captionof{table}{Comparison with previous perception-free planners.}
  \label{tab:data_scale}
  \setlength{\tabcolsep}{4pt}
  \begin{tabular}{l|cccc}
    \toprule
     & \textbf{LAW} & \textbf{World4Drive} & \textbf{Epona} & \textbf{Ours} \\
    \midrule
    Encoder size & 21M & 21M & \textbf{1.1B} & 307M \\
    Data scale & \textasciitilde20h & \textasciitilde20h & 128h & \textbf{330h} \\
    PDMS & 83.8 & 85.1 & 86.1 & \textbf{89.0} \\
    \bottomrule
  \end{tabular}
  \vspace{-8pt}
\end{wrapfigure}

\paragraph{Driving Video Dataset Curation and Scaling} We initialize the ViT encoder with parameters released by V-JEPA 2 \cite{assran2025v}. To bridge the domain gap, we curate a large-scale driving video dataset from three publicly available datasets: CoVLA \cite{arai2025covla}, DrivingDojo \cite{wang2024drivingdojo}, and OpenScene(trainval split) \cite{sima2023_occnet}. All videos are captured from a front-view camera and processed into 8-frame clips at a resolution of $512\times256$ and 2 Hz. We adopt the V-JEPA objective to train the ViT encoder in a self-supervised manner on this curated dataset. As shown in Table~\ref{tab:data_scale}, thanks to the efficiency of the latent prediction task and effective mode-collapse prevention, we successfully scale pretraining to 330 hours with lower computational cost than prior methods.

\paragraph{Perception-free End-to-End Autonomous Driving}
Following prior world-model-based end-to-end driving works, we adopt a perception-free setting for evaluation, where the model is supervised solely by human trajectories without relying on perception annotations \cite{lienhancing}. Given spatiotemporal features extracted by the ViT encoder from the front-view image, future waypoints are predicted using a transformer decoder with learnable queries. Given the front-view inputs $I_t^1$ and $I_{t-1}^1$, we extract spatiotemporal features using the pretrained ViT encoder and denote them as $\mathbf{F}_t \in \mathbb{R}^{N_f \times D}$. We introduce $M$ learnable query embeddings $\mathbf{Q} \in \mathbb{R}^{M \times D}$, each corresponding to one future waypoint. The transformer-based decoder \cite{vaswani2017attention} attends to $\mathbf{F}_t$ via cross-attention to produce
\[
\mathbf{H} = \mathrm{TransformerDecoder}(\mathbf{Q}, \mathbf{F}_t),
\]
which is then mapped to predicted waypoints $\hat{W}_t = \mathrm{MLP}(\mathbf{H})$, where $\hat{W}_t=\{\hat{w}_t^1,\ldots,\hat{w}_t^M\}$ and each $\hat{w}_t^i = (\hat{x}_t^i, \hat{y}_t^i, \hat{\psi}_t^i)$ represents the BEV position and heading at time step $t+i$. The network is trained end-to-end using a MSE loss between $\hat{W}_t$ and the ground-truth trajectory $W_t$. Despite its simplicity, this setup significantly outperforms prior methods (Table~\ref{tab:data_scale}), highlighting the effectiveness of V-JEPA-based driving video pretraining.

\subsection{Waypoint-anchored Proposals Generation}

Building upon the strong representations from Driving Video Pretraining, we design a planner that follows a proposal-selection paradigm. As mentioned before, a fixed vocabulary can be seen as proposals but suffers from discretization error. Inspired by iPad \cite{guo2025ipad}, we instead generate proposals online.

Given the visual features $\mathbf{F}_t \in \mathbb{R}^{N_f \times D}$ and ego status at time $t$, we project the ego status by a linear layer into an ego feature $\mathbf{e}_t \in \mathbb{R}^{1\times D}$. The proposal queries are initialized as $\mathbf{Q}_0\in\mathbb{R}^{N_p\times M\times D}$ by adding $\mathbf{e}_t$ to learnable positional embeddings, where $N_p$ is the number of waypoint-trajectory proposals and $M$ is the number of future waypoints. We iteratively refine the proposal queries $\mathbf{Q}_{\ell}$ for $L$ iterations. At iteration $\ell$, an MLP decodes $\mathbf{Q}_{\ell}$ into waypoint-trajectory proposals $\tilde{W}_{\ell}=\{\tilde{W}_{\ell}^{(n)}\}_{n=1}^{N_p}$, with $\tilde{W}_{\ell}\in\mathbb{R}^{N_p\times M\times 3}$ and each waypoint $(x,y,\psi)$. Using these explicit waypoint locations as anchors, we refine the queries by exchanging information among proposals and aggregating features from $\mathbf{F}_t$ around each predicted waypoint via lift-splat BEV feature sampling \cite{philion2020lift}. We then update the queries with a lightweight MLP:
\[
\mathbf{Q}_{\ell+1} = \mathrm{MLP}\!\left(\mathrm{WADA}(\mathbf{Q}_{\ell}, \tilde{W}_{\ell}, \mathbf{F}_t)\right),
\]
where WADA denotes Waypoint-anchored Deformable Attention \cite{xia2022vision}.

Because the final trajectory for planning is selected from $\tilde{W}_{L}$, its distribution is critical. Given a human trajectory $W_t$ and the intermediate proposals $\{\tilde{W}_{\ell}\}_{\ell=0}^{L-1}$, a naive way to guide $\tilde{W}_{\ell}$ is using the minimum-over-$N$ loss \cite{gupta2018social} with discounted supervision across iterations:

\[
\mathcal{L}_{traj}=\sum_{\ell=0}^{L-1}\lambda^{L-\ell-1}\min_{n\in\{1,\dots,N_p\}}\left\lVert W_t-\tilde{W}_{\ell}^{(n)}\right\rVert_2,
\]

where $\lambda = 0.1$ down-weights earlier iterations to encourage coarse-to-fine refinement.

However, in autonomous driving, there are often multiple valid choices beyond the single human trajectory for a scene. This naive guidance method limits the multimodality of the proposals. We present our solution in the next section.

\subsection{Multimodal Trajectories Distillation} \label{sec:MTD}

To alleviate sparse supervision from a single human trajectory per scene, we distill knowledge from rule-based simulators. HydraMDP \cite{li2024hydra} performs hydra-distillation by learning scores over a fixed vocabulary. Instead, we let the simulator provide multimodal trajectory targets to guide the proposal distribution.

\begin{wrapfigure}{r}{0.42\textwidth}
    \centering
    \vspace{-10pt}
    \includegraphics[width=0.95\linewidth]{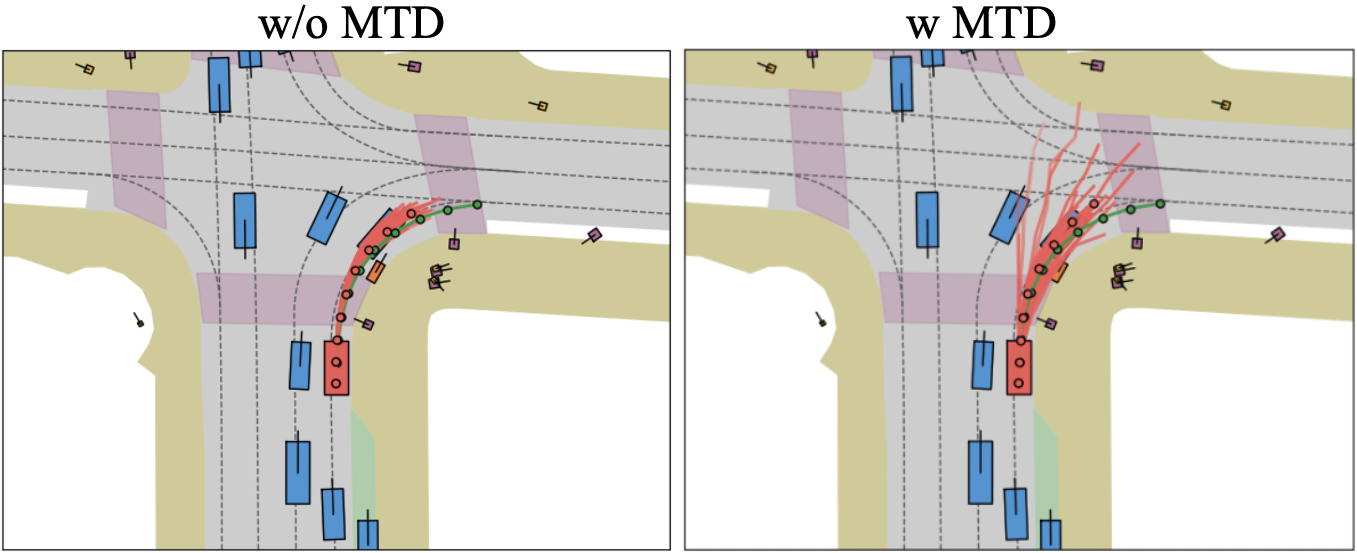}
    \caption{Bird's eye view of proposals.}
    \label{fig:demo_proposals}
    \vspace{-10pt}
\end{wrapfigure}
Concretely, we start by building a trajectory vocabulary following VADv2 and HydraMDP, but use the vocabulary for a different purpose. We gather all trajectories in the training dataset, which includes more than 100k trajectories. Then we use a clustering method, k-means \cite{douze2024faiss}, to select trajectory centers. We select 8192 centers as the trajectory vocabulary, balancing coverage and computational cost. For each scene in the training dataset, we select high quality multimodal trajectories from the vocabulary using rule-based simulators. Following NAVSIM v2 \cite{cao2025pseudo, caesar2021nuplan}, we calculate the EPDM score for all trajectories. We refer to the Appendix \ref{sec:metrics} for the detailed definition. Specifically, we first run a PID controller to convert 8 waypoints into a denser 41-point trajectory. Then, at each timestep, we replay other road agents, traffic lights, etc., and compute collisions and other metrics. The rule-based simulator in NAVSIM v2 is designed only for evaluation. We further improve the vectorized computation efficiency to meet large-scale offline scoring needs. After obtaining scores for all trajectories in the vocabulary across all scenes in the training dataset, we select a group of multimodal trajectories $\mathcal{P}_t=\{P_t^{1},\ldots,P_t^{N_{pseudo}}\}$ for each scene by ranking and thresholding. During training, we use these multimodal trajectories  as pseudo-teachers to guide the proposals, instead of a single human trajectory. The final $\mathcal{L}_{traj}$ is defined as:

\begin{equation}
\begin{split}
 \sum_{\ell=1}^{L}\lambda^{L - \ell}\Big( \min \lVert W_t - \tilde{W}_{\ell}^{(n)} \rVert_2
 + \sum_{P \in \mathcal{P}_t}\min \lVert P - \tilde{W}_{\ell}^{(n)} \rVert_2 \Big).
\end{split}
\end{equation}

\noindent
Here, the $\min$ operator takes the minimum over the proposal index $n \in \{1,\ldots,N_p\}$ at iteration $\ell$. As shown in Figure \ref{fig:demo_proposals}, without Multimodal Trajectories Distillation (MTD), the proposals exhibit clear mode collapse; with MTD, they become multimodal.

\subsection{Momentum-aware Trajectory Selection}

To select the best trajectory among the proposal set for planning, we train a neural scorer to evaluate the final proposals $\{\tilde{W}_{L}^{(n)}\}_{n=1}^{N_p}$. Concretely, we apply max pooling over the waypoint dimension to the proposal queries $\mathbf{Q}_{L} \in \mathbb{R}^{N_p \times M \times D}$ to obtain pooled features $\bar{\mathbf{Q}}_{L} \in \mathbb{R}^{N_p \times D}$, which are then fed into a multi-layer perceptron (MLP) to produce proposal scores $S \in \mathbb{R}^{N_p \times 1}$. The scorer is trained with a binary cross-entropy (BCE) loss:
\[
\mathcal{L}_{\text{score}}=\mathrm{BCE}(S,\hat{S}),
\]
where $\mathrm{BCE}(x,y)=-y\log x-(1-y)\log(1-x)$. The supervision $\hat{S}$ is derived from simulator-based EPDMS evaluation and is also used to define candidate pseudo-targets $\mathcal{P}_t=\{P_t^{1},\ldots,P_t^{N_{pseudo}}\}$.

While Multimodal Trajectory Distillation improves proposal diversity, it can amplify temporal inconsistency, increasing discomfort due to larger variation across adjacent time frames. To mitigate this, we make the score momentum-aware by incorporating a comfort term. Let $\hat{W}_{t-1}$ denote the selected trajectory at the previous time frame. We compute a distortion-based comfort score $S_c \in \mathbb{R}^{N_p \times 1}$ by comparing $\hat{W}_{t-1}$ with each current proposal in $\{\tilde{W}_{L}^{(n)}\}_{n=1}^{N_p}$, and recalibrate the learned score via
\[
S \leftarrow \frac{7\,S + \,S_c}{8}.
\]
where the weights are following NAVSIM v2.
Finally, the selected trajectory is formalized as
\[
\hat{W}_t = \tilde{W}_{L}^{(n^{*})}, \quad n^{*}=\arg\max_{n\in\{1,\ldots,N_p\}} S_n,
\]
where $S_n$ denotes the recalibrated score of proposal $\tilde{W}_{L}^{(n)}$.

\subsection{Losses}

In end-to-end driving tasks, it is important to add auxiliary tasks to enhance the model's environment understanding capability, e.g., BEV map segmentation, 3D object detection, and tracking. However, these traditional dense understanding tasks are computationally intensive. Here we use the lightweight auxiliary tasks instead \cite{guo2025ipad}, which contain rich spatiotemporal signals and are compatible with the proposal-centric design. 

We use two auxiliary tasks: proposal-centric mapping and collision prediction. For the first task, our model predicts the on-road and on-route probabilities of the proposed waypoints in $\tilde{W}_{\ell}$, denoted as $R \in \mathbb{R}^{N_p \times M \times 2}$. The proposal-centric mapping loss is $\mathcal{L}_{map} = \mathrm{BCE}(R, \hat{R})$. For proposal-centric collision prediction, we estimate the collision probability $A_v$ of waypoints in $\tilde{W}_{\ell}$ using log-replay simulation. This task not only requires the model to detect surrounding objects, but also to understand their moving pattern. The proposal-centric collision loss is $\mathcal{L}_{colli} = \lVert A_v - \hat{A}_v \rVert + 0.1 \, \mathrm{BCE}(A_v, \hat{A}_v)$.

Our Drive-JEPA is end-to-end differentiable. The training loss is defined as:
\[
\mathcal{L} = \mathcal{L}_{traj} + w_{score}\mathcal{L}_{score} + w_{map}\mathcal{L}_{map} + w_{colli}\mathcal{L}_{colli},
\]
where $w_{score} = 1$, $w_{map} = 2$ and $w_{colli} = 1$.
\section{Experiments} \label{sec:experiments}

\subsection{Dataset and Metrics}

We evaluate our method on three benchmarks, including NAVSIM v1, NAVSIM v2 and Bench2Drive.

\paragraph{NAVSIM v1} NAVSIM is a real-world dataset based on OpenScene \cite{sima2023_occnet} and NuPlan \cite{caesar2021nuplan}. It contains 103k and 12k diverse and challenging driving scenarios for model training (Navtrain) and evaluation (Navtest), and introduces simulation-based metrics to better review closed-loop planning capability through open-loop evaluation. During evaluation, the output trajectory is evaluated by a simulator to get rule-based simulation metric scores, including No at-fault Collisions(NC), Drivable Area Compliance(DAC), Time to Collision with bounds(TTC), Ego Progress(EP) and Comfort(C). The final PDM Score(PDMS) is derived by aggregating these metrics:

\begin{equation}
PDMS = NC \times DAC \times \frac{5 \times (EP + TTC) + 2 \times C}{12}
\end{equation}

\paragraph{NAVSIM v2} Compared with NAVSIM v1, NAVSIM v2 strengthens driving-quality evaluation by extending PDMS to EPDMS with richer rule-compliance and comfort assessment. It adds Driving Direction Compliance (DDC), Traffic Light Compliance (TLC), and Lane Keeping (LK) to better capture traffic-rule adherence, and replaces the original comfort term with History Comfort (HC) and Extended Comfort (EC) to evaluate both short- and longer-horizon smoothness.

\paragraph{Bench2Drive} Bench2Drive \cite{jia2024bench} is a closed-loop evaluation benchmark based on CARLA \cite{dosovitskiy2017carla}, designed to assess end-to-end autonomous driving systems in interactive urban scenarios. The evaluation includes 220 routes spanning 44 diverse, interactive scenarios. Official metrics include Driving Score (DS), Success Rate (SR), Efficiency and Comfortness, which collectively measure navigation performance, safety, and rule adherence. For detailed metric definitions, see Appendix \ref{sec:metrics}.

\subsection{Implementation Details}

In the Driving Video Pretraining stage, we use 8 H800 GPUs and train for 50 epochs, which takes about 3 days. The Drive-JEPA planners are trained on two NVIDIA A30 GPUs for 20 epochs with a total batch size of 64, using the Adam \cite{kingma2014adam} optimizer with a learning rate of \(1 \times 10^{-4}\), while the ViT encoder uses a learning rate of \(1 \times 10^{-5}\). We set \(N_p = 32\) proposals, which is efficient while achieving strong performance in our ablation studies. We use only the front-view camera, resized to \(512 \times 256\). Despite using fewer input images than prior methods and no LiDAR, we outperform them. See Appendix \ref{sec:details} for details.

\begin{table*}[th]
\caption{Quantitative comparisons on NAVSIM v1. \textbf{Bold}/\underline{underlined} mark the best/second-best; the first block is the perception-free setting.}
\label{tab:navsim_v1_results}
\centering
\setlength{\tabcolsep}{3pt}
\renewcommand{\arraystretch}{0.95}
\begin{tabular}{l|c|c|cc|ccc|G}
\toprule
\textbf{Method} & \textbf{Backbone} & \textbf{Inputs} & \textbf{NC\,$\uparrow$} & \textbf{DAC\,$\uparrow$} & \textbf{EP\,$\uparrow$} & \textbf{C\,$\uparrow$} & \textbf{TTC\,$\uparrow$} & \textbf{PDMS\,$\uparrow$} \\ \midrule
LAW \cite{lienhancing}  &   resnet34   & C \& L  &  97.4  &  93.3   & 78.8 &  \textbf{100}   &  91.9 &  83.8   \\ 
World4Drive \cite{zheng2025world4drive}  &   resnet34   & C \& L  &  97.4  &  94.3  &  79.9  &  \textbf{100}   & 92.8 &  85.1   \\ 
Epona \cite{zhang2025epona}  &   ViT/G   & Camera  &  \underline{97.9}  &  \underline{95.1}  & \underline{80.4}  & 99.9 & \underline{93.8}  &  \underline{86.2}   \\ 
Ours &     ViT/L     &  Camera  &  \textbf{98.7}  &  \textbf{96.2}  & \textbf{82.9}  & \textbf{100} & \textbf{95.5} &  \textbf{89.0}   \\  \midrule
Transfuser \cite{chitta2022transfuser}   &  ResNet34   &    C \& L    &  97.7  &  92.8  &  79.2  &  \textbf{100}   & 92.8 & 84.0   \\
HydraMDP \cite{li2024hydra}   &  ResNet34   &  C \& L  & 98.3 & 96.0 & 78.7 & \textbf{100} & 94.6 & 86.5   \\ 
HydraMDP++ \cite{li2025hydra}   &  ResNet34   &  C \& L  &  97.6 & 96.0 & 80.4 & \textbf{100} & 93.1 & 86.6   \\ 
DiffusionDrive \cite{liao2025diffusiondrive}   &  ResNet34   &  C \& L  &  98.2  & 96.2 & 82.2 & \textbf{100} & 94.7 & 88.1   \\ 
GoalFLow  \cite{xing2025goalflow}  &  ResNet34   &  C \& L  &  \underline{98.4}  & \textbf{98.3} & 85.0 & \textbf{100} & 94.6 & 90.3   \\ 
DriveDPO \cite{shang2025drivedpo}   &  ResNet34   &  C \& L  &  \textbf{98.5}  & \underline{98.1} & 84.3 & \textbf{100} & \underline{94.8} & 90.0   \\ 
DriveSuprim \cite{yao2025drivesuprim}   &  ResNet34   &  Camera  &  97.8  & 97.3 & 86.7 & \textbf{100} & 93.6 & 89.9   \\ 
iPad \cite{guo2025ipad}   &  ResNet34   &  Camera  &  \underline{98.4}  & 97.9 & \underline{87.4} & 99.9 & \textbf{94.9} & \underline{91.1}   \\ 
Drive-JEPA(Ours)   &  ResNet34   &  Camera  &  98.2  & 98.0 & \textbf{88.8} & 99.9 & 94.2 & \textbf{91.5}   \\   \midrule
Hydra-MDP \cite{li2024hydra}   &  ViT/L  & C \& L  &  98.4  &  97.7 &  85.0 & \textbf{100} & 94.5 &  89.9    \\
iPad \cite{guo2025ipad}   &  ViT/L   &  Camera  &  \textbf{99.2}  & 97.4 & 87.8 & 99.7 & \textbf{96.3} & 91.7   \\ 
DriveSuprim \cite{yao2025drivesuprim}  &  ViT/L   &  Camera  &  98.6  & \textbf{98.6} & \underline{91.3} & \textbf{100} & 95.5 & \underline{93.5}   \\
DrivR \cite{kirby2026drivor}  &  ViT/L   &  Camera  &  98.9  & \underline{98.3} & 89.1 & \textbf{100} & \underline{96.2} & 93.1   \\
Drive-JEPA(Ours)   &  ViT/L  & Camera  & \underline{99.1} & 98.2 &  \textbf{91.4} & 99.9 & \underline{96.2} &  \textbf{93.7}    \\  \bottomrule
\end{tabular}
\end{table*}

\begin{table*}[h]
\caption{Quantitative comparisons on NAVSIM v2.}
\label{tab:navsim_v2_results}
\centering
\setlength{\tabcolsep}{3pt}
\begin{tabular}{l|c|cccc|ccccc|G}
\toprule
\textbf{Method} & \textbf{Backbone} & \textbf{NC$\uparrow$} & \textbf{DAC$\uparrow$} & \textbf{DDC$\uparrow$} & \textbf{TL$\uparrow$} & \textbf{EP$\uparrow$} & \textbf{TTC$\uparrow$} & \textbf{LK$\uparrow$} & \textbf{HC$\uparrow$} & \textbf{EC$\uparrow$} & \textbf{EPDMS$\uparrow$} \\ \midrule
Transfuser   &  ResNet34  & 96.9 & 89.9 & 97.8 & 99.7 & \underline{87.1} & 95.4 & 92.7 & \textbf{98.3} & \textbf{87.2} & 76.7    \\
HydraMDP++ & ResNet34 & 97.2 & \underline{97.5} & \textbf{99.4} & 99.6 & 83.1 & 96.5 & 94.4 & 98.2 & 70.9 & 81.4 \\
DriveSuprim & ResNet34 & 97.5 & 96.5 & \textbf{99.4} & 99.6 & \textbf{88.4} & 96.6 & 95.5 & \textbf{98.3} & 77.0 & 83.1 \\
iPad   &  ResNet34  & \underline{98.7} & \textbf{97.8} & 99.1 & \textbf{99.8} & 84.0 & \textbf{98.0} & \underline{96.0} & 98.0 & 68.2 & \underline{84.1}    \\
Ours    &  ResNet34  &  \textbf{98.8} & 97.4 & 99.0 & \textbf{99.8} & 83.5 & \textbf{98.0} & \textbf{96.2} & 98.1 & \underline{85.6} & \textbf{85.4}   \\  \midrule
HydraMDP++   &  ViT/L  &  \underline{98.5} & 98.5 & \underline{99.5} & 99.7 & 87.4 & 97.9 & 95.8 & 98.2 & 75.7 & 85.6    \\
iPad &  ViT/L  &  \textbf{98.7} & 98.0 & 98.9 & \textbf{99.8} & 86.6 & \textbf{98.3} & \underline{97.2} & \textbf{98.3} & 74.6 & 85.8   \\
DriveSuprim & ViT/L & 98.4 & \textbf{98.6} & \textbf{99.6} & \textbf{99.8} & 90.5 & \underline{97.8} & 97.0 & \textbf{98.3} & \underline{78.6} &  \underline{87.1} \\
Ours  &  ViT/L  &   98.4  &  \textbf{98.6} &  99.1 & \textbf{99.8} & 88.4 & \underline{97.8} &  \textbf{97.6} &  97.9 & \textbf{84.8} & \textbf{87.8}    \\ \bottomrule
\end{tabular}
\end{table*}

\subsection{Main Results}

\myparagraph{Results on NAVSIM v1} As shown in Table \ref{tab:navsim_v1_results}, Drive-JEPA achieves the best PDMS with both ResNet34 and ViT/L backbones, surpassing the strongest prior method DriveSuprim \cite{yao2025drivesuprim} by 0.2 PDMS under ViT/L. Notably, while maintaining high safety metrics such as NC and TTC, our method achieves the best DAC and Ego Progress, resulting in safer yet more assertive driving.

\myparagraph{Perception-free End-to-end Autonomous Driving} We also evaluated our method in a perception-free setting, where we use a simple decoder with the pretrained ViT encoder as described in \ref{subsec:video_pretraining}. As shown in Table \ref{tab:navsim_v1_results}, our method surpasses previous methods by a large margin, regardless of backbone size. The PDMS is even close to SOTA methods that rely on perception annotations, demonstrating the strength of V-JEPA pretraining.

\myparagraph{Results on NAVSIM v2} NAVSIM v2 has more sophisticated metrics than NAVSIM v1. Our method still outperformances all prior methods. While prior methods struggle with EC, our method performs quite well while achieving good results on safety metrics, traffic rule compliance and Ego Progress.

\begin{table}[h]
\centering
\begin{minipage}[t]{0.52\textwidth}
  \centering
  \footnotesize
  \caption{Quantitative comparisons on Bench2Drive.}
  \label{tab:bench2drive_results}
  \vspace{2pt}
  \begin{tabular}{c|cccc}
    \toprule
   \textbf{Method} & \textbf{Effi.\,$\uparrow$} & \textbf{Comf.\,$\uparrow$} & \textbf{SR\,$\uparrow$} & \textbf{DS\,$\uparrow$} \\ \midrule
     AD-MLP & 48.45 & 22.63 & 0.00 & 18.05 \\
     UniAD & 129.21 & \underline{43.58} & 16.36 & 45.81 \\
     VAD & \underline{157.94} & \textbf{46.01} & 15.00 & 42.35 \\
     TCP & 76.54 & 18.08 & 30.00 & 59.90 \\
     DriveDPO & \textbf{166.80} & 26.79 & 30.62 & 62.02 \\
     iPad & 153.83 & 35.51 & 33.18 & 60.52 \\ %% eval_pad_b2d_baseline_2/merged.json
     DriveTransformer & 100.64 & 20.78 & \underline{35.01} & \underline{63.46} \\
     Ours & 157.85 & 30.24 & \textbf{36.82} & \textbf{64.52} \\ %% eval_pad_pseudo_teacher_a2/merged.json
    \bottomrule
  \end{tabular}
\end{minipage}\hfill
\begin{minipage}[t]{0.45\textwidth}
  \centering
  \footnotesize
  \setlength{\tabcolsep}{3pt}
  \caption{Comparison with mainstream vision pretraining methods.}
  \label{tab:compare_encoder}
  \vspace{2pt}
  \begin{tabular}{l|c|G}
    \toprule
    \textbf{Vision Encoder} & \textbf{Size} & \textbf{PDMS\,$\uparrow$} \\ \midrule
    Epona \cite{zhang2025epona}              & ViT/G    & \underline{86.2} \\ \midrule
    ImageNet \cite{deng2009imagenet}         & ResNet34 & 76.0 \\
    DepthAnything \cite{yang2024depth}       & ViT/L    & - \\
    MAE \cite{he2022masked}                  & ViT/L    & - \\
    Dinov2 \cite{oquab2023dinov2}            & ViT/L    & 76.1 \\
    Sigclip \cite{zhai2023sigmoid}           & ViT/L    & 83.4 \\
    V-JEPA 2 \cite{assran2025v}              & ViT/L    & 86.1 \\ \midrule
    Ours                                     & ViT/L    & \textbf{89.0} \\
    \bottomrule
  \end{tabular}
\end{minipage}
\end{table}

\myparagraph{Results on Bench2Drive} Bench2Drive evaluates autonomous agents in close-loop simulation. Our method achieves the best Driving Score with very competitive Efficiency. Our method surpasses another proposal-centric method iPad by 4 in Driving Score, identifying the effectiveness of Multimodal Trajectories Distillation.

\newcommand{\cmark}{\textcolor{black}{\Checkmark}}
\newcommand{\xmark}{\textcolor{black}{\XSolidBrush}}

\begin{table*}[t]
\caption{Ablation study of the proposed modules on NAVSIM v2.}
\label{tab:ablation}
\centering
\setlength{\tabcolsep}{3pt}
\begin{tabular}{cccc|cccc|ccccc|c|g}
\toprule 
\textbf{$\mathcal{M}_1$} & \textbf{$\mathcal{M}_2$} & \textbf{$\mathcal{M}_3$} & \textbf{$\mathcal{M}_4$} & \textbf{NC$\uparrow$} & \textbf{DAC$\uparrow$} & \textbf{DDC$\uparrow$} & \textbf{TL$\uparrow$} & \textbf{EP$\uparrow$} & \textbf{TTC$\uparrow$} & \textbf{LK$\uparrow$} & \textbf{HC$\uparrow$} & \textbf{EC$\uparrow$} & \textbf{$\mathcal{D}$$\uparrow$} & \textbf{EPDMS$\uparrow$} \\ \midrule
\xmark & \xmark &  \xmark  & \xmark &  98.7 & 97.8 & 99.1 & 99.8 & 84.0 & 98.0 & 96.0 & 98.0 & 68.2 & 25\% & 84.1     \\ 
\cmark & \xmark &  \xmark  & \xmark &  98.7 & 98.0 & 98.9 & 99.8 & 86.6 & 98.3 & 97.2 & 98.3 & 74.6 & 21\% & $85.8_{\scriptsize \textcolor{green!50!black}{+1.7}}$   \\ 
\xmark & \cmark &  \xmark  & \xmark &  98.3  &  98.1 &  99.1 & 99.9 & 89.1 &  97.7 &  97.7 & 98.1 & 69.7 & 24\% & $86.1_{\scriptsize \textcolor{green!50!black}{+2.0}}$    \\ 
\xmark & \cmark &  \cmark  & \xmark &  98.5  &  98.6 &  99.1 & 99.8 & 89.1 &  97.9 &  97.6 & 97.8 & 47.9 & 40\% & $84.5_{\scriptsize \textcolor{green!50!black}{+0.4}}$    \\ 
\xmark & \cmark &  \cmark  & \cmark &  98.4  &  98.6 &  99.1 & 99.8 & 88.4 & 97.8 &  97.6 &  97.9 & 84.8 & 40\% & $87.8_{\scriptsize \textcolor{green!50!black}{+3.7}}$    \\ \bottomrule
\end{tabular}
\end{table*}

\begin{figure*}[h]
  \centering
  \includegraphics[width=1.0\textwidth, trim=0 2cm 0 0, clip]{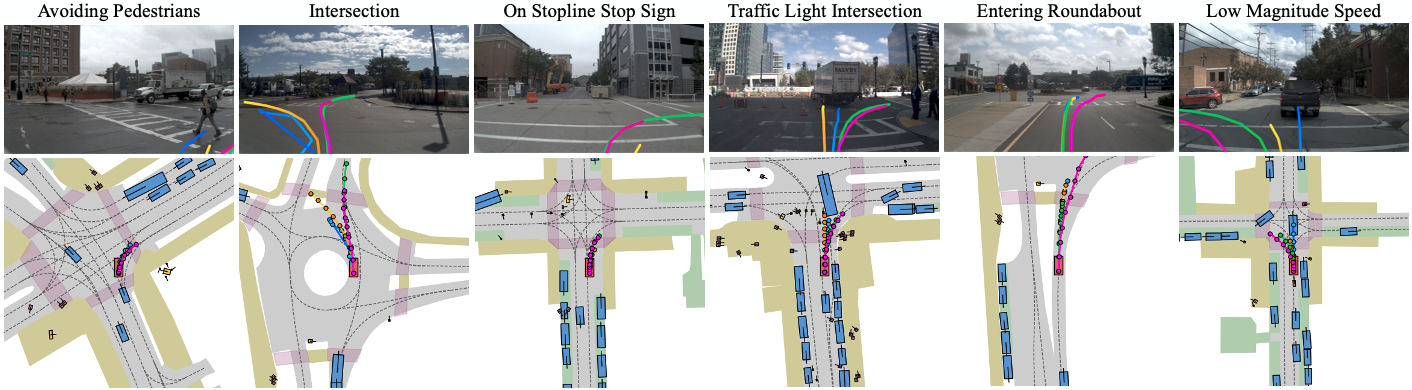}
  \caption{Qualitative comparison of trajectories by different models in front-facing camera and bird's eye view on different driving scenarios. Trajectories are shown for:
  \protect\tikz[baseline=-0.5ex]{\protect\shade[left color=green!70!black, right color=green!50!yellow] (0,-0.05) rectangle (0.4,0.05); \protect\fill[green!70!black] (0.2,0) circle (2.5pt);} Human Trajectory,
  \protect\tikz[baseline=-0.5ex]{\protect\shade[left color=magenta, right color=magenta!70!red] (0,-0.05) rectangle (0.4,0.05); \protect\fill[magenta] (0.2,0) circle (2.5pt);} Drive-JEPA,
  \protect\tikz[baseline=-0.5ex]{\protect\shade[left color=cyan!80!blue, right color=cyan!50!blue] (0,-0.05) rectangle (0.4,0.05); \protect\fill[cyan!80!blue] (0.2,0) circle (2.5pt);} iPad,
  \protect\tikz[baseline=-0.5ex]{\protect\shade[left color=orange, right color=yellow] (0,-0.05) rectangle (0.4,0.05); \protect\fill[orange] (0.2,0) circle (2.5pt);} Transfuser.
  }
  \label{fig:planning_demos}
\end{figure*}

\subsection{Ablation Studies}

\begin{figure}[h]
\centering
\begin{minipage}[c]{0.50\textwidth}
  \centering
  \footnotesize
  \captionof{table}{Ablation on number of pseudo-teacher trajectories.}
  \label{tab:num_pseudo_t}
  \vspace{2pt}
  \begin{tabular}{c|ccccc}
    \toprule
     $N_{pseudo}$ & 0 & 1 & 2 & 4 & 8  \\ \midrule
     EPDMS & 87.2 & 87.8 & 87.7 & 87.8 & 87.5 \\
    \bottomrule
  \end{tabular}
\end{minipage}\hfill
\begin{minipage}[c]{0.45\textwidth}
  \centering
  \includegraphics[width=0.95\linewidth]{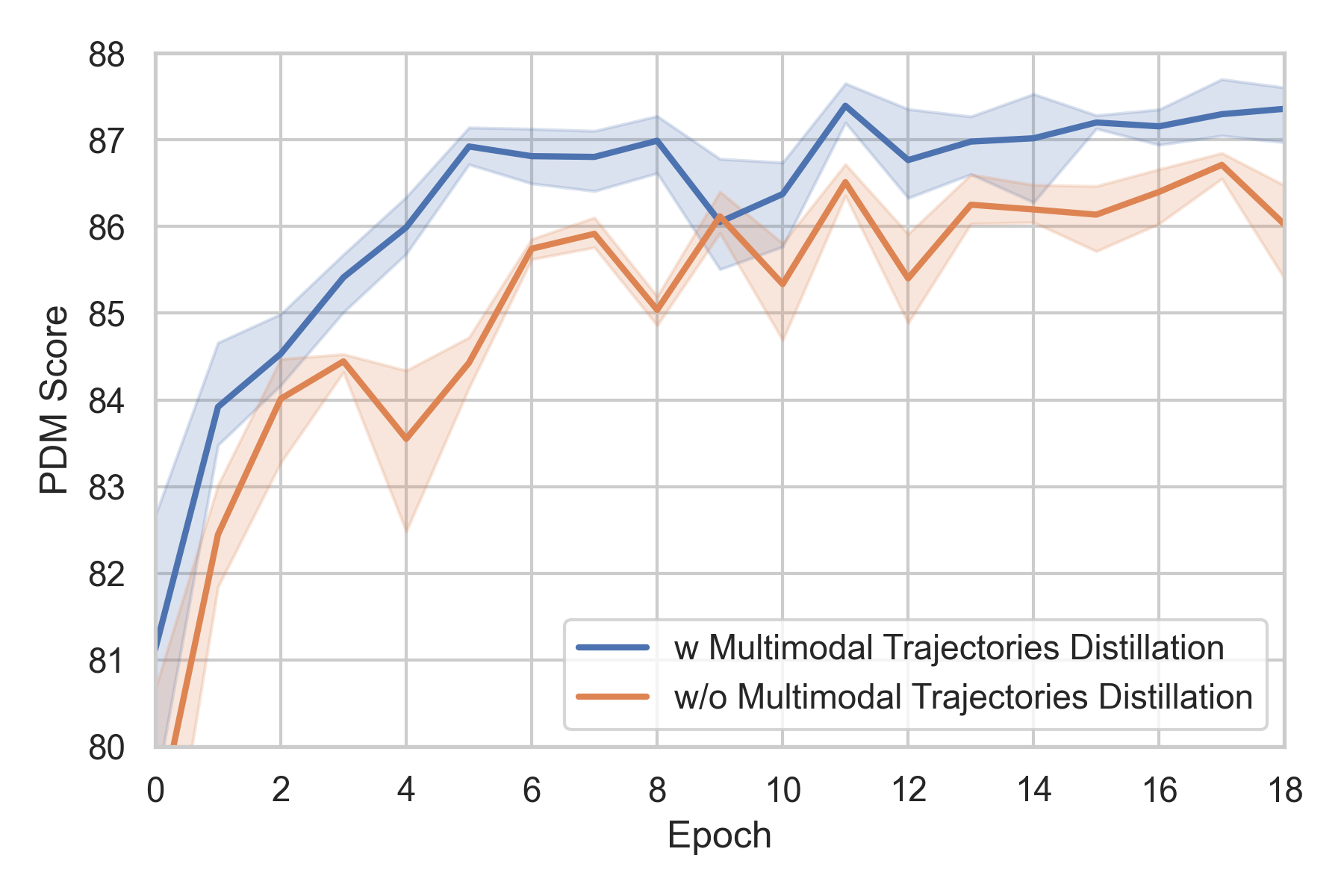}
  \caption{Multimodal Trajectory Distillation improves PDM score.}
  \label{fig:shaded_plot}
\end{minipage}
\end{figure}

\myparagraph{Ablation studies on proposed modules} We first conducted ablation studies on the proposed modules: $\mathcal{M}_1$: V-JEPA 2 checkpoints, $\mathcal{M}_2$: Driving Video Pretraining, $\mathcal{M}_3$: Multimodal Trajectories Distillation, and $\mathcal{M}_4$: Momentum-aware Trajectory Selection. As shown in Table \ref{tab:ablation}, replacing ResNet34 with the ViT released by V-JEPA 2 ($\mathcal{M}_1$) improves EPDMS. Our Driving Video Pretraining further boosts performance by reducing the domain gap ($\mathcal{M}_2$). After adding $\mathcal{M}_3$, the framework achieves better $\mathcal{D}$ (Diversity) \cite{liao2025diffusiondrive} and overall metrics, which is also supported by the validation score curve in Figure \ref{fig:shaded_plot}. However, the increased diversity results in worse EC. Finally, adding $\mathcal{M}_4$ not only largely boosts EC to 84.8, but also sets a new best record on EPDMS.

\myparagraph{Ablation study on the number of Pseudo Teacher Trajectories.} As shown in Table \ref{tab:num_pseudo_t}, we tried \( N_{pseudo} = 0, 1, 2, 4, 8 \) pseudo-teacher trajectories. While the correlation between \( N_{pseudo} \) and EPDMS is not very strong, using pseudo-teacher trajectories consistently performs better than using none (\( N_{pseudo}=0 \)).

\myparagraph{Ablation on Driving Video Pretraining.} We use the same simple decoder with encoders pretrained by mainstream pretraining methods. As shown in Table \ref{tab:compare_encoder}, V-JEPA 2 performs the best among them. MAE and DepthAnything could not converge. This highlights the strength of the V-JEPA objective for video pretraining. In this work, we curated a large driving video dataset. The ViT/L encoder trained on this dataset with the V-JEPA objective further boosts performance, surpassing the SOTA Epona by 3 PDMS.
\FloatBarrier
\section{Conclusion}

We proposed Drive-JEPA, an end-to-end driving framework that combines V-JEPA video pretraining with multimodal trajectory distillation to mitigate imitation-learning modal collapse. Pretraining a ViT encoder on large-scale driving videos yields strong planning representations, enabling a simple decoder to achieve competitive perception-free performance. Distilling simulator-guided pseudo-teacher trajectories improves proposal diversity, and momentum-aware selection further enhances temporal stability and comfort. Drive-JEPA achieves state-of-the-art results on NAVSIM v1/v2 and improves closed-loop performance on Bench2Drive.

% Auto-hidden during submission; fill in for camera-ready.
\begin{ack}
% Funding and acknowledgments go here for the camera-ready version.
\end{ack}

\bibliographystyle{plainnat}
\bibliography{refs}

\appendix
\section{Metrics} \label{sec:metrics}

\subsection{Extended Predictive Driver Model Score (EPDMS)} 

NAVSIM v2~\cite{cao2025pseudo} extends the PDMS metric from NAVSIM v1 to EPDMS, which is defined as:
\begin{equation}
EPDMS = NC \times DAC \times DDC \times TLC \times \frac{5 \times (EP + TTC) + 2 \times (LK + HC + EC)}{16}
\end{equation}

The subscores include: No at-fault Collision (NC), Drivable Area Compliance (DAC), Driving Direction Compliance (DDC), Traffic Light Compliance (TLC), Ego Progress (EP), Time to Collision (TTC), Lane Keeping (LK), History Comfort (HC), and Extended Comfort (EC).

Among these, DDC, TLC, LK, HC, and EC are newly introduced in NAVSIM v2. We summarize them below; for full computation details, we refer readers to NAVSIM v2~\cite{cao2025pseudo}.

\paragraph{Driving Direction Compliance (DDC).} The ego vehicle must follow the legal direction of travel within lanes and avoid driving in oncoming lanes outside intersections.

\paragraph{Traffic Light Compliance (TLC).} This score evaluates whether the ego vehicle obeys traffic-light phases and enters intersections only under a valid green signal.

\paragraph{Lane Keeping (LK).} This score measures whether the ego vehicle stays near the centerline of the current lane and avoids lingering between adjacent lanes, while discouraging hesitant “half-commit” lane-change probes.

\paragraph{History Comfort (HC).} To better assess ride comfort, we prepend the predicted trajectory with a short segment of the human driver’s recent motion using a fixed padding length of 1.5 seconds. The resulting continuous trajectory is then evaluated using the same comfort metric adopted in the nuPlan framework~\cite{caesar2021nuplan}.

\paragraph{Extended Comfort (EC).} This score checks that the predicted motion remains smooth across consecutive time steps.

\subsection{Diversity ($\mathcal{D}$)} 

We report this metric in Table~\ref{tab:ablation}. Following DiffusionDrive~\cite{liao2025diffusiondrive}, it is defined as:
\[
\mathcal{D}
= 1 - \frac{1}{N_p}\sum_{i=1}^{N_p}
\frac{\mathrm{Area}\!\left(\tilde{W}_{t_i}\,\cap\,\bigcup_{j=1}^{N_p}\tilde{W}_{t_j}\right)}
{\mathrm{Area}\!\left(\tilde{W}_{t_i}\,\cup\,\bigcup_{j=1}^{N_p}\tilde{W}_{t_j}\right)} \, .
\]

To compute the IoU between trajectories, we rasterize each trajectory polyline into a 2D occupancy mask by buffering the polyline with a 2-meter width (i.e., a 2\,m-thick corridor) and projecting it onto a grid.

\subsection{Metrics used in Bench2Drive}

We briefly describe the four metrics used in Bench2Drive; for full computation details, we refer readers to Bench2Drive~\cite{jia2024bench}.

\paragraph{Success Rate (SR).} The proportion of routes completed successfully within the allotted time and without traffic violations.

\paragraph{Driving Score (DS).} This score follows the official CARLA~\cite{dosovitskiy2017carla} metric, combining route completion with penalties for infractions.

\paragraph{Efficiency.} CARLA includes a check for excessively low speed by comparing the ego vehicle’s speed with nearby traffic.

\paragraph{Comfort.} This metric follows nuPlan’s~\cite{caesar2021nuplan} smoothness (comfort) protocol, which evaluates longitudinal acceleration (min/max), the maximum absolute lateral acceleration, yaw rate, yaw acceleration, longitudinal jerk, and the maximum magnitude of the jerk vector.

\section{More details} \label{sec:details}

\paragraph{Threshold.} In Section~\ref{sec:MTD}, we use a threshold on simulated EPDMS to select trajectories from the vocabulary. The threshold is set to 0.95. In many scenes, more than $N_{\text{pseudo}}$ trajectories exceed this threshold; during training, we uniformly sample $N_{\text{pseudo}}$ trajectories at random from this high-quality subset.

\begin{table}[ht]
\centering
\caption{Input image resolution.}
\label{tab:resolution}
\begin{tabular}{lc}
\toprule
Method & Input image resolution \\
\midrule
Transfuser & 1024$\times$256 \\
HydraMDP++ & 1024$\times$256 \\
DriveSuprim & 1024$\times$256 \\
GoalFlow & 1024$\times$256 \\
iPad & 4$\times$768$\times$432 \\
Ours & 2$\times$512$\times$256 \\
\bottomrule
\end{tabular}
\end{table}

\paragraph{Resolution.} As shown in Table~\ref{tab:resolution}, HydraMDP++, DriveSuprim, and GoalFlow follow Transfuser in using an input resolution of $1024\times256$, formed by stacking the front, left, and right camera images. iPad uses a higher resolution ($768\times432$) and four camera views (front, left, right, and back). Our setting uses only the front camera at $512\times256$. We include both $I_t$ and $I_{t-1}$, resulting in an input tensor of $2\times512\times256$.

\section{More visualization}

\begin{figure}[h]
    \centering
    \vspace{-2pt}
\includegraphics[width=1.0\textwidth]{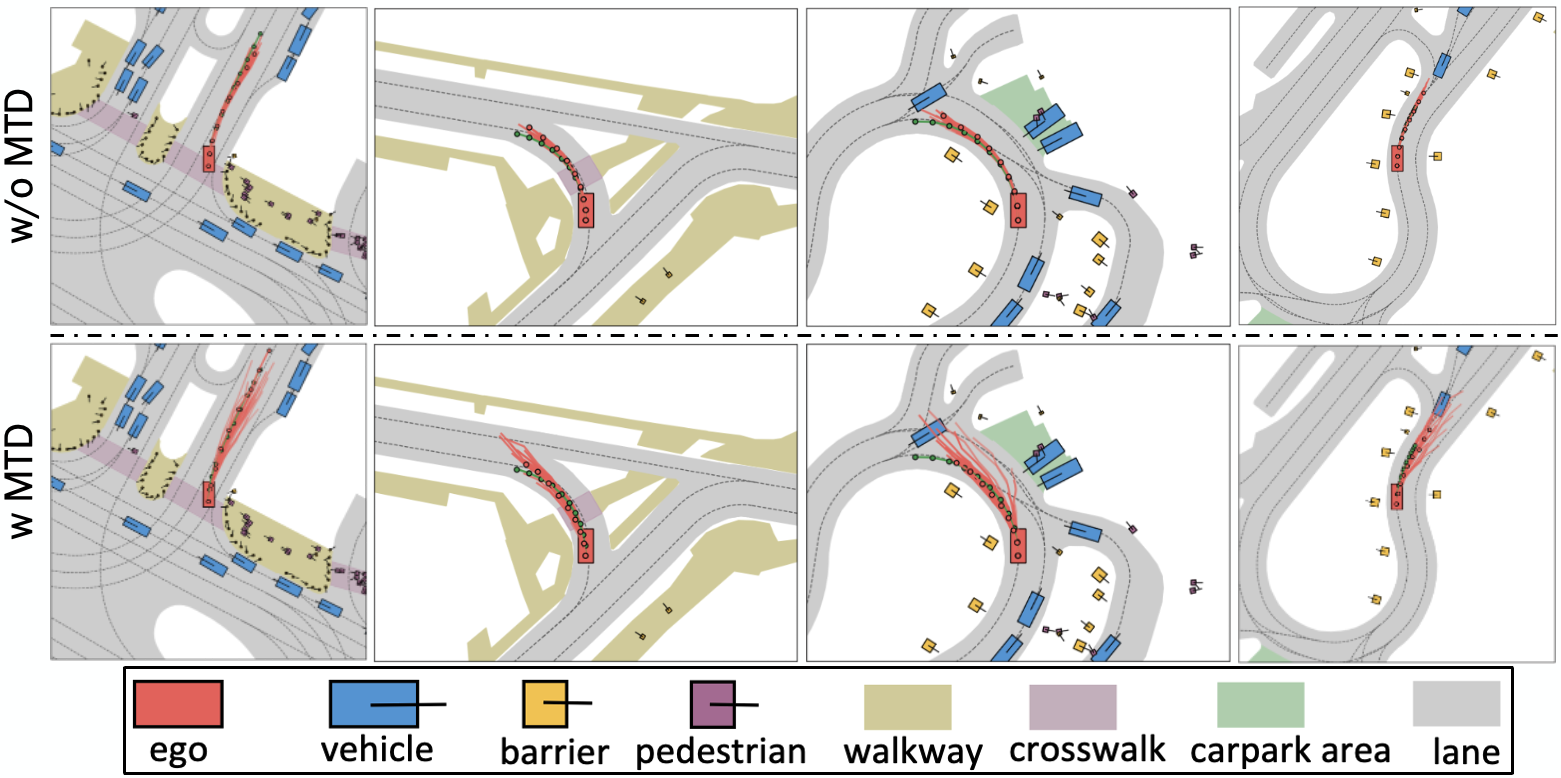}
    \caption{Bird's eye view of proposals. Without Mutimodal Trajectory Distillation (MTD), the proposals collapse into one mode. With MTD, the proposals show multimodal distribution.}
    \label{fig:more_demo_proposals}
    \vspace{-2pt}
\end{figure}

\newpage
\section*{NeurIPS Paper Checklist}

\begin{enumerate}

\item {\bf Claims}
    \item[] Question: Do the main claims made in the abstract and introduction accurately reflect the paper's contributions and scope?
    \item[] Answer: \answerYes{}
    \item[] Justification: The abstract and Section~\ref{sec:intro} state our three contributions (V-JEPA driving pretraining, multimodal trajectory distillation, and momentum-aware trajectory selection); the reported 93.3 PDMS on NAVSIM v1 and 87.8 EPDMS on NAVSIM v2 are supported by the experimental results in Section~\ref{sec:experiments}.

\item {\bf Limitations}
    \item[] Question: Does the paper discuss the limitations of the work performed by the authors?
    % TODO: Add a "Limitations" paragraph in 5_conclusion.tex (single-front-camera assumption,
    % simulator pseudo-teacher dependence, compute requirements) and flip this answer to \answerYes{}.
    \item[] Answer: \answerNo{}
    \item[] Justification: The current draft does not include a dedicated limitations discussion; we plan to add one covering the single-front-camera assumption, the dependence on simulator-generated pseudo-teacher trajectories, and computational requirements before camera-ready.

\item {\bf Theory assumptions and proofs}
    \item[] Question: For each theoretical result, does the paper provide the full set of assumptions and a complete (and correct) proof?
    \item[] Answer: \answerNA{}
    \item[] Justification: The paper does not include formal theoretical results or proofs; the contributions are empirical (a pretraining recipe, a planner, and a selection module).

    \item {\bf Experimental result reproducibility}
    \item[] Question: Does the paper fully disclose all the information needed to reproduce the main experimental results of the paper to the extent that it affects the main claims and/or conclusions of the paper (regardless of whether the code and data are provided or not)?
    \item[] Answer: \answerYes{}
    \item[] Justification: Section~\ref{sec:method} fully describes the architecture, the V-JEPA pretraining objective, the proposal generator, and the momentum-aware selection module; Section~\ref{sec:experiments} and Appendix~\ref{sec:details} list datasets, splits, optimizer, batch size, hardware, input resolution, and the EPDMS pseudo-teacher threshold needed to reproduce the main results.

\item {\bf Open access to data and code}
    \item[] Question: Does the paper provide open access to the data and code, with sufficient instructions to faithfully reproduce the main experimental results, as described in supplemental material?
    \item[] Answer: \answerNo{}
    \item[] Justification: All datasets used (NAVSIM v1, NAVSIM v2, and Bench2Drive) are publicly released. We do not provide an anonymized code link at submission time, but we will release the full code and pretrained checkpoints upon acceptance.

\item {\bf Experimental setting/details}
    \item[] Question: Does the paper specify all the training and test details (e.g., data splits, hyperparameters, how they were chosen, type of optimizer) necessary to understand the results?
    \item[] Answer: \answerYes{}
    \item[] Justification: Section~\ref{sec:experiments} and Appendix~\ref{sec:details} specify the dataset splits, input resolution, optimizer, learning rate, batch size, number of proposals~$N_p$, the EPDMS pseudo-teacher threshold of 0.95, and the evaluation protocol used.

\item {\bf Experiment statistical significance}
    \item[] Question: Does the paper report error bars suitably and correctly defined or other appropriate information about the statistical significance of the experiments?
    % TODO: Either rerun the main experiments with multiple seeds and report std-dev, or keep this
    % answer as \answerNo{} with the justification below.
    \item[] Answer: \answerNo{}
    \item[] Justification: We follow the convention in NAVSIM and Bench2Drive evaluation (e.g., DriveSuprim, iPad, GoalFlow), reporting single-run scores; multi-seed runs of the full pipeline are computationally prohibitive given the cost of V-JEPA pretraining and proposal-centric training.

\item {\bf Experiments compute resources}
    \item[] Question: For each experiment, does the paper provide sufficient information on the computer resources (type of compute workers, memory, time of execution) needed to reproduce the experiments?
    \item[] Answer: \answerYes{}
    \item[] Justification: Compute resources are reported in the experiment details (Section~\ref{sec:experiments}): driving video pretraining uses 8 NVIDIA H800 GPUs for 50 epochs (\textasciitilde 3 days), and the Drive-JEPA planners are trained on 2 NVIDIA A30 GPUs for 20 epochs with a total batch size of 64.

\item {\bf Code of ethics}
    \item[] Question: Does the research conducted in the paper conform, in every respect, with the NeurIPS Code of Ethics \url{https://neurips.cc/public/EthicsGuidelines}?
    \item[] Answer: \answerYes{}
    \item[] Justification: We have reviewed the NeurIPS Code of Ethics; the work uses only publicly available driving datasets, does not involve human subjects, and does not release any high-risk artifact.

\item {\bf Broader impacts}
    \item[] Question: Does the paper discuss both potential positive societal impacts and negative societal impacts of the work performed?
    % TODO: Expand the justification below into a 2-3 sentence broader-impacts paragraph (positive:
    % planning safety/comfort; negative: simulator-pseudo-label over-reliance, real-world validation
    % needed) -- ideally place it in 5_conclusion.tex or as a brief Broader Impact paragraph there.
    \item[] Answer: \answerYes{}
    \item[] Justification: Positive: more reliable end-to-end planners can improve driving safety and comfort. Negative: over-reliance on simulator-generated pseudo-teacher trajectories may transfer simulator artifacts to deployment; real-world validation beyond NAVSIM/Bench2Drive is required before any deployed system.

\item {\bf Safeguards}
    \item[] Question: Does the paper describe safeguards that have been put in place for responsible release of data or models that have a high risk for misuse (e.g., pre-trained language models, image generators, or scraped datasets)?
    \item[] Answer: \answerNA{}
    \item[] Justification: The paper does not release a high-risk artifact; the released model is a driving planner trained on public driving data, with no generative output that could be misused.

\item {\bf Licenses for existing assets}
    \item[] Question: Are the creators or original owners of assets (e.g., code, data, models), used in the paper, properly credited and are the license and terms of use explicitly mentioned and properly respected?
    % TODO: Verify and list license strings for each asset (NAVSIM v1/v2, Bench2Drive, OpenScene,
    % nuPlan, CoVLA/DrivingDojo if used, V-JEPA codebase) before submission.
    \item[] Answer: \answerYes{}
    \item[] Justification: All datasets and codebases used (NAVSIM v1/v2, Bench2Drive, nuPlan, V-JEPA, and the driving-video corpora used for pretraining) are cited at first use in Sections~\ref{sec:method}--\ref{sec:experiments} with their original papers; their licenses and terms of use are respected.

\item {\bf New assets}
    \item[] Question: Are new assets introduced in the paper well documented and is the documentation provided alongside the assets?
    \item[] Answer: \answerNA{}
    \item[] Justification: The paper does not release a new dataset; the curated driving-video corpus consists entirely of previously released public datasets used under their original licenses, with no derivative dataset published.

\item {\bf Crowdsourcing and research with human subjects}
    \item[] Question: For crowdsourcing experiments and research with human subjects, does the paper include the full text of instructions given to participants and screenshots, if applicable, as well as details about compensation (if any)?
    \item[] Answer: \answerNA{}
    \item[] Justification: The paper does not involve crowdsourcing or research with human subjects.

\item {\bf Institutional review board (IRB) approvals or equivalent for research with human subjects}
    \item[] Question: Does the paper describe potential risks incurred by study participants, whether such risks were disclosed to the subjects, and whether Institutional Review Board (IRB) approvals (or an equivalent approval/review based on the requirements of your country or institution) were obtained?
    \item[] Answer: \answerNA{}
    \item[] Justification: The paper does not involve human-subjects research, so no IRB review was required.

\item {\bf Declaration of LLM usage}
    \item[] Question: Does the paper describe the usage of LLMs if it is an important, original, or non-standard component of the core methods in this research? Note that if the LLM is used only for writing, editing, or formatting purposes and does \emph{not} impact the core methodology, scientific rigor, or originality of the research, declaration is not required.
    %this research?
    % TODO: Confirm LLMs were used at most for writing/editing assistance. If LLMs were used as a
    % core methodological component, switch to \answerYes{} and describe the usage.
    \item[] Answer: \answerNA{}
    \item[] Justification: LLMs were not used as an important, original, or non-standard component of the core method; any LLM use was limited to incidental writing or editing assistance, which the policy does not require declaring.

\end{enumerate}

\end{document}